%% file: main.tex
\documentclass{article}
\usepackage{iclr2020_conference,times,hyperref,url,amsfonts,graphicx}
\newcommand{\circled}[1]{\textcircled{\tiny{\fontfamily{cmss}\selectfont #1}}}
\iclrfinalcopy

\title{The perceptual boost of visual attention is task-dependent in naturalistic settings}
\author{Freddie Bickford Smith, Xiaoliang Luo, Brett D Roads, Bradley C Love\\University College London\\\texttt{\{f.bickfordsmith.18,xiao.luo.17,b.roads,b.love\}@ucl.ac.uk}}

\begin{document}
\maketitle
\input{sections/abstract.tex}
\input{sections/introduction.tex}
\input{sections/method.tex}
\input{sections/results.tex}
\input{sections/discussion.tex}
\input{sections/conclusion.tex}
\input{sections/acknowledgements}
\input{sections/references.tex}
\end{document}

%% file: sections/abstract.tex
\begin{abstract}
Top-down attention allows people to focus on task-relevant visual information. Is the resulting perceptual boost task-dependent in naturalistic settings? We aim to answer this with a large-scale computational experiment. First, we design a collection of visual tasks, each consisting of classifying images from a chosen task set (subset of ImageNet categories). The nature of a task is determined by which categories are included in the task set. Second, on each task we train an attention-augmented neural network and then compare its accuracy to that of a baseline network. We show that the perceptual boost of attention is stronger with increasing task-set difficulty, weaker with increasing task-set size and weaker with increasing perceptual similarity within a task set.
\end{abstract}

%% file: sections/introduction.tex
\section{Introduction}
\label{sec:intro}
By modulating neural representations of visual stimuli, top-down attention enables prioritised processing of task-relevant information. Past work has shown that this mechanism enhances people's perceptual abilities and that its influence varies between visual tasks \citep{carrasco}. Does this hold for a broad range of naturalistic tasks, like recognising an object in an everyday scene? To our knowledge, the literature lacks a systematic characterisation of this issue. At the same time, thanks to faster computers and larger sets of annotated data, it has become straightforward to train deep convolutional neural networks to perform such visual tasks \citep{krizhevsky}. Analyses have shown that these networks are state-of-the-art computational models of the human visual system \citep{yamins}. Hence there is a gap in the visual-attention literature, and there is a computational-modelling technique that has only recently become practically usable and validated by neuroscience. Our work addresses the gap with the modelling technique. By studying the performance of neural networks on a number of visual tasks, we seek to establish how the perceptual boost of attention varies with the nature of a task.

Each task we consider consists of classifying images from a chosen \textit{task set}, a subset of categories from the ImageNet dataset \citep{russakovsky}. We define three quantitative dimensions along which a task set can vary: difficulty, size and perceptual similarity (see Section \ref{sec:mthd-properties} for definitions). Our hypothesis is that each of these \textit{task-set properties} is important in determining the perceptual boost of attention on a task. We base this on observations and intuitions. First, ImageNet categories are known to range in difficulty, likely due to variation in factors such as object scale, image clutter and shape distinctiveness \citep{russakovsky}. If attention interacts with such factors, its impact on perception could be expected to vary with task-set difficulty. Second, increasing the breadth of a task (eg, performing classification on examples from a larger number of categories) generally introduces more severe tradeoffs when learning to weight the features of neural representations. If this is true, and if attention is framed as modulating representations, the size of a task set should influence how effectively attention can act. Third, defining attention as before, the similarity of representations within a task set might affect how usefully attention can shape them.

Our experiment\footnote{Code and data are available at \url{https://github.com/fbickfordsmith/attention-iclr}.} is based on a collection of task sets that vary with respect to the task-set properties we define. For each task set, we construct an attention-augmented neural network (more concisely, an \textit{attention network}) by taking an ImageNet-pretrained VGG16 \citep{simonyan}, fixing its weights and inserting a trainable attention mechanism. Then we train the attention network solely on images from the task set. The perceptual boost of attention is measured by comparing the accuracy of this optimised attention network to the accuracy of a baseline network. In this way, we gather evidence to assess how attention's impact on perception varies with task-set properties.

%% file: sections/method.tex
\section{Method}
\subsection{Task-set properties}
\label{sec:mthd-properties}
Let task set $\mathcal{C}$ be a subset of the 1000 image categories in the ImageNet dataset. We define three task-set properties: difficulty, size and perceptual similarity. By choosing which categories to include in a task set, we control its properties. The \textit{difficulty} of $\mathcal{C}$ is the mean error rate of an ImageNet-pretrained VGG16 on categories in $\mathcal{C}$:
\begin{equation}
\textrm{difficulty}(\mathcal{C}) =
\frac{1}{|\mathcal{C}|}
\sum_{c_i \in \mathcal{C}}
(1 - \textrm{accuracy}_{\textsc{vgg}}(c_i))
\end{equation}
where accuracy takes values between 0 and 1. The \textit{size} of $\mathcal{C}$ is the number of categories it contains: 
\begin{equation}
\textrm{size}(\mathcal{C}) =
|\mathcal{C}|
\end{equation}
The \textit{perceptual similarity} of $\mathcal{C}$ is the mean pairwise similarity of the categories in it:
\begin{equation}
\textrm{similarity}(\mathcal{C}) =
\frac{1}{|\mathcal{C}|^2 - |\mathcal{C}|}
\sum_{c_i \in \mathcal{C}}\sum_{c_j \in \mathcal{C}}
\mathbb{I}(i \neq j) s(c_i,c_j)
\end{equation}
where $\mathbb{I}$ is an indicator function and $s(c_i,c_j)$ is the cosine similarity between the average representations of categories $c_i$ and $c_j$. That is,
\begin{equation}
s(c_i,c_j) =
\frac{r(c_i) \cdot r(c_j)}{\|r(c_i)\| \|r(c_j)\|} \qquad
r(c) = \frac{1}{|\mathcal{X}_c|}
\sum_{x_i \in \mathcal{X}_c} \textrm{VGG}'(x_i) \qquad
r \in \mathbb{R}^{4096}
\end{equation}
where $\mathcal{X}_c$ is the subset of the training images for which the label is $c$, and $\textrm{VGG}'$ denotes VGG16 with its final layer removed.

\subsection{Neural networks}
\label{sec:mthd-models}
We use an ImageNet-pretrained VGG16 as a foundational neural network, and incorporate visual attention as a linear modulation of neural representations \citep{lindsay}. VGG16 computes a probability distribution over the 1000 possible states of the category label, $c$, for a given image, $x$:
\begin{equation}
p(c|x) = \textrm{VGG}(x)
\end{equation}
The network can be decomposed into two parts. The convolutional layers, $\textrm{VGG}_1$, transform $x$ to a latent representation, $z$. The fully-connected layers, $\textrm{VGG}_2$, transform $z$ to $p(c|x)$. That is,
\begin{equation}
\textrm{VGG}_1: x \rightarrow z \qquad
\textrm{VGG}_2: z \rightarrow p(c|x) \qquad
x \in \mathbb{R}^{224 \times 224 \times 3} \qquad
z \in \mathbb{R}^{7 \times 7 \times 512}
\end{equation}
We define attention to be the multiplication of $z$ by nonnegative attention weights, $a$. Our attention network computes
\begin{equation}
p(c|x,a) = \textrm{VGG}_2(a \odot \textrm{VGG}_1(x)) \qquad
a \in \mathbb{R}_{\geq 0}^{7\times7\times512}
\end{equation}
where $\odot$ denotes an elementwise multiplication. We treat the pretrained $\textrm{VGG}_1$ and $\textrm{VGG}_2$ as fixed functions. The attention weights are the only trainable parameters and are initialised to 1.

\subsection{Experiment design}
Using the task-set properties defined in Section \ref{sec:mthd-properties}, we assemble three groups of task sets. \textit{Difficulty-based} task sets vary substantially in difficulty but not in size or perceptual similarity. \textit{Size-based} task sets are diverse in size but similar in difficulty and perceptual similarity. \textit{Similarity-based} task sets have a wide range of perceptual-similarity values but constant size and near-constant difficulty.

\begin{figure}[h]
\centering
\vspace{5pt}
\includegraphics[width=0.75\textwidth, trim={0 0.4cm 0 0.2cm}]{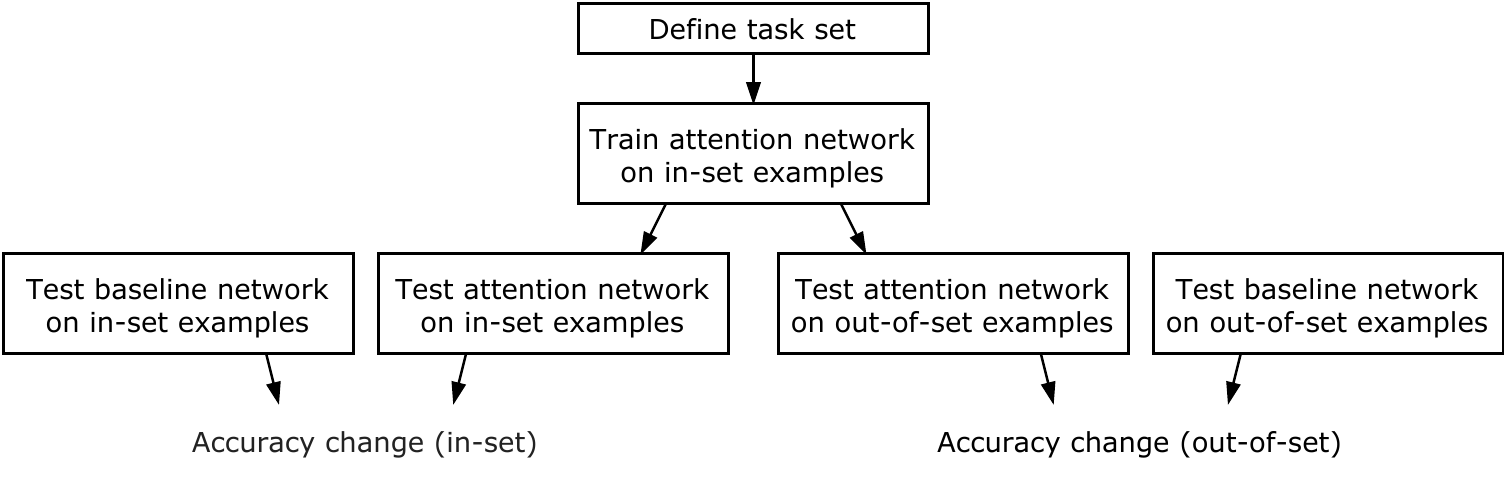}
\caption{
    Procedure for measuring how the nature of an image-classification task, controlled by the choice of task set, affects the accuracy change produced by attention. This is performed for each task set.}
\label{fig:exp-design}
\end{figure}

For each task set, we construct a new attention network and train it solely on images from the task set. Then we compare the accuracy of this attention network to the accuracy of a baseline network (a task-agnostic attention network trained on all ImageNet categories). We make this comparison on examples both from within the task set (giving the change in \textit{in-set} accuracy) and from outside the set (giving the change in \textit{out-of-set} accuracy). Our primary interest is the perceptual boost that attention produces on the task for which it is optimised (ie, the change in in-set accuracy), but for completeness we also include results for out-of-set accuracy. Figure \ref{fig:exp-design} summarises our approach.

%% file: sections/results.tex
\section{Results}
\label{sec:results}
\begin{figure}[ht]
\centering
\includegraphics[width=0.8\textwidth, trim={0 0.9cm 0 0.2cm}]{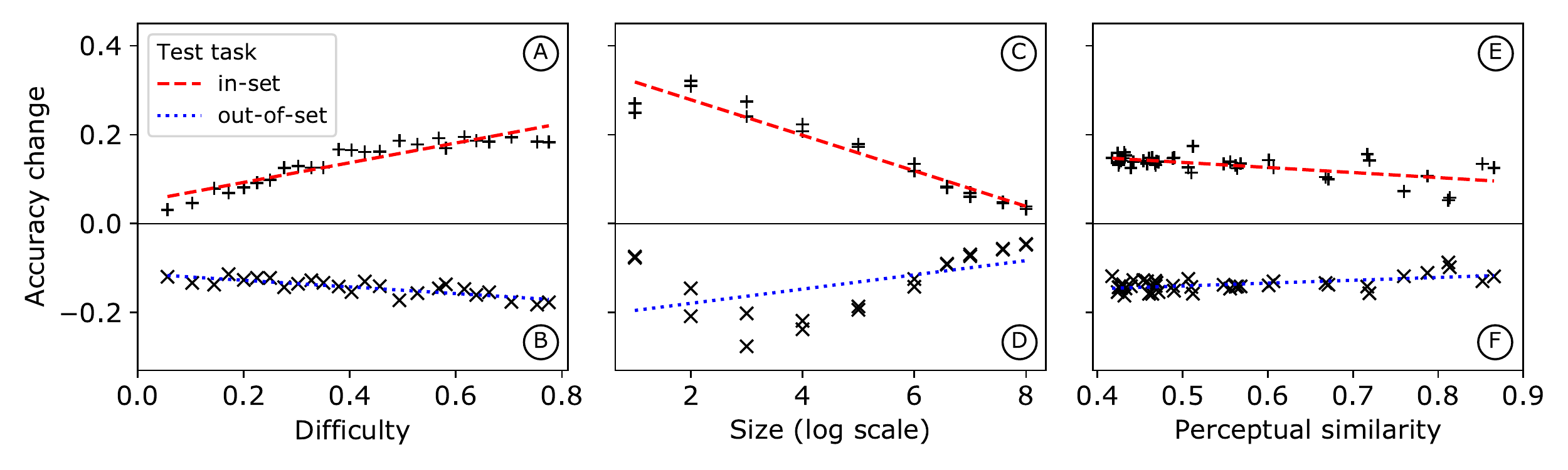}
\caption{
    Accuracy change produced by attention on 25 difficulty-based task sets (left), 20 size-based task sets (middle) and 40 similarity-based task sets (right). Task-set size is transformed logarithmically with base 2. Least-squares linear regression is applied to each subset of results, from \circled{A} to \circled{F}; predictions of the linear models are shown as broken lines.}
\label{fig:plots}
\end{figure}

For all tasks, applying attention results in higher in-set accuracy and lower out-of-set accuracy (see Figure \ref{fig:plots} and Table \ref{tab:stats}). This effect is stronger with increasing task-set difficulty, weaker with increasing task-set size and weaker with increasing perceptual similarity within a task set.

\begin{table}[ht]
{\small
\vspace{5pt}
\centering
\begin{tabular}{rllrrrrr}
\hline
  & Property & Test task & \multicolumn{1}{c}{$\rho$} & \multicolumn{1}{c}{$\tau_b$} & \multicolumn{1}{c}{$\beta_0$} & \multicolumn{1}{c}{$\beta_1$} & \multicolumn{1}{c}{$R^2$} \\
\hline
    \circled{A} & difficulty & in-set & 0.93 & 0.79 & 0.05 & 0.22 & 0.85 \\
    \circled{B} & difficulty & out-of-set & -0.83 & -0.65 & -0.11 & -0.07 & 0.68 \\
    \circled{C} & size & in-set & -0.97 & -0.91 & 0.36 & -0.04 & 0.93 \\
    \circled{D} & size & out-of-set & 0.69 &  0.59 &  -0.21 & (0.02) &  0.27 \\
    \circled{E} & similarity & in-set & -0.53 & -0.38 & 0.19 & -0.11 & 0.39 \\
    \circled{F} & similarity & out-of-set & (0.37) & (0.27) & -0.17 & 0.06 &  0.31 \\
\hline
\end{tabular}
\caption{
    Relationships between the three task-set properties defined in Section \ref{sec:mthd-properties} (the controlled experimental variables) and the accuracy change produced by attention (the measured experimental variable). Five statistics are shown for each subset of results, from \circled{A} to \circled{F}: Spearman's $\rho$; Kendall's $\tau_b$; coefficient of determination, $R^2$; regression intercept, $\beta_0$; regression slope, $\beta_1$. For each value of $\rho$, $\tau_b$, $\beta_0$ and $\beta_1$, except those in parentheses, $p<0.001$.}
\label{tab:stats}
}
\end{table}

%% file: sections/discussion.tex
\section{Discussion}
The task-set properties defined in Section \ref{sec:mthd-properties} have significant influence on the perceptual boost of visual attention (see Section \ref{sec:results}). First, the more difficult a task set is, the greater the impact of attention on in-set accuracy (the performance measure we focus on in this discussion). This is perhaps because attention mitigates the effects of factors that contribute towards difficulty (eg, object scale, image clutter and shape distinctiveness). Second, increasing the number of categories in a task set results in a weaker attention-induced boost to in-set accuracy. The mechanism underlying this might be to do with the weight tradeoffs involved when training a neural network \citep{sutton}. A visual feature might be discriminative for images from some categories but confounding for images from others; the weight placed on that feature is determined by its net contribution to accuracy across the whole training set. Optimising weights on a restricted set of categories (all tasks sets in our experiment contain at most 256 categories) corresponds to a relaxation of this weight tradeoff, allowing the network to make use of features that previously might have been dominated by others. (Similar logic might explain the difficulty result: high-difficulty categories are those whose most indicative features are most severely drowned out in the training process; specialising on fewer categories has an outsized effect for more difficult task sets.) Third, the greater the perceptual similarity of image categories within a task set, the less pronounced the impact of attention on performance. This could be because images from similar categories are represented alike in neural activations (this is how we define perceptual similarity) and a linear reweighting by attention does not much enhance the network's ability to discriminate between them.

%% file: sections/conclusion.tex
\section{Conclusion}
The perceptual boost of visual attention varies substantially between naturalistic tasks. As well as informing further basic research on attention, this finding has practical implications for neural-network designers deciding whether to use a visual-attention mechanism. Empirically testing our suggested explanations of the results would be an interesting direction for future work. So would a study of the perceptual boost of attention when there is covariance between task-set properties (our experiment was designed to minimise this).

%% file: sections/acknowledgements.tex
\section*{Acknowledgements}
We thank Edward Grefenstette for helpful discussions. This work was supported by NIH Grant 1P01HD080679, Wellcome Trust Investigator Award WT106931MA and Royal Society Wolfson Fellowship 183029 to Bradley C Love.